\def\year{2020}\relax
\documentclass[letterpaper]{article} 
\usepackage{aaai20}  
\usepackage{times}  
\usepackage{helvet} 
\usepackage{courier}  
\usepackage[hyphens]{url}  
\usepackage{graphicx} 
\usepackage{graphicx}  
\usepackage{amsmath}
\usepackage{amsthm}
\usepackage{amssymb}

\usepackage{algorithm}
\usepackage[noend]{algorithmic}

\usepackage{multirow}
\usepackage{tabularx}

\usepackage{microtype}

\usepackage{subcaption}
\usepackage{mathtools}

\def\rot{\rotatebox}

\newtheorem{definition}{Definition}

\title{An Efficient Explorative Sampling Considering the Generative Boundaries of \\Deep Generative Neural Networks}
\author{
Giyoung Jeon\thanks{Equal Contribution}, Haedong Jeong\footnotemark[1]\\
Ulsan National Institute of Science and Technology\\
50, UNIST-gil, Ulsan 44919, Republic of Korea\\
\{giyoung, haedong\}@unist.ac.kr
\And Jaesik Choi\thanks{Corresponding Author}\\
Korea Advanced Institute of Science and Technology\\
291 Daehak-ro, Daejeon 34141, Republic of Korea\\
jaesik.choi@kaist.ac.kr
}

\begin{document}

\maketitle
\begin{abstract}
Deep generative neural networks (DGNNs) have achieved realistic and high-quality data generation. In particular, the adversarial training scheme has been applied to many DGNNs and has exhibited powerful performance. Despite of recent advances in generative networks, identifying the image generation mechanism still remains challenging.
In this paper, we present an explorative sampling algorithm to analyze generation mechanism of DGNNs. Our method efficiently obtains samples with identical attributes from a query image in a perspective of the trained model. We define generative boundaries which determine the activation of nodes in the internal layer and probe inside the model with this information.
To handle a large number of boundaries, we obtain the essential set of boundaries using optimization. By gathering samples within the region surrounded by generative boundaries, we can empirically reveal the characteristics of the internal layers of DGNNs. We also demonstrate that our algorithm can find more homogeneous, the model specific samples compared to the variations of $\epsilon$-based sampling method.
\end{abstract}

\begin{figure*}[t!]
    \centering
    \begin{subfigure}{0.98\textwidth}        \includegraphics[width=\textwidth]{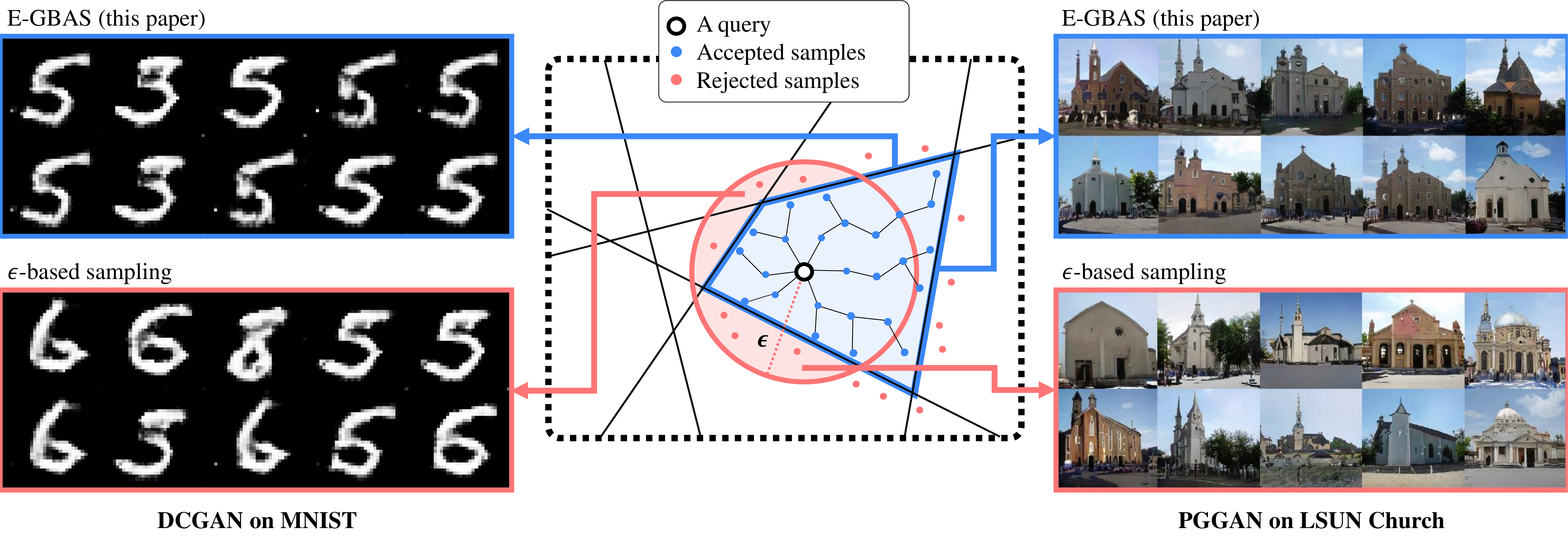}
    \end{subfigure}
    \caption{Illustrative examples of our explorative generative boundary aware sampling (E-GBAS) and $\epsilon_{L_2}$-based sampling.}\label{fig:main_fig}
\end{figure*}

\section{Introduction}
The primary objective of a generative model is to generate realistic data. Recently proposed adversarial training schemes, such as generative adversarial networks (GANs), have exhibited remarkable performance not only in terms of the quality of each instance but also the diversity of the generated data. Despite those improvements, the generation mechanism inside the generative models is not well-studied.

In general, a generative model maps a point in the latent space to a sample in the data space. In other words, data instances are embedded as latent vectors in a perspective of the trained generative model. A latent space is divided by boundaries derived from the structure of the model, where the vectors in the space represent the generation information according to which side of boundaries they are placed. We utilize these characteristics to examine the generation mechanism of the model.

When we select an internal layer and a latent vector in the DGNNs, there exists the corresponding region which is established by a set of boundaries. Samples in this region have the same activation pattern and deliver similar generation information to the next layer. The details of the delivered information can be identified indirectly by comparing the generated outputs from these samples. Given a DGNN trained to generate human faces, for example, if we identify the region in which samples share a certain hair color but vary in others characteristics (eye, mouth, etc.), such a region would be related to the generation of the same hair color.

However, it is non-trivial to obtain samples from the region with desired properties of DGNNs because (1) thousands of generative boundaries are involved in the generation mechanism and (2) a linear modification in the input dimension may cause a highly non-linear change in the internal units and the output. Visiting the previous example again, there may exist regions with different hair colors, distinct attributes, or their combinations. Furthermore, a small linear modification of the vector in the latent space may change the entire output \cite{szegedy2013intriguing}. To overcome this difficulty, an efficient algorithm to identify the appropriate region and explore the space are necessary.

In this paper, we propose an efficient, explorative sampling algorithm to reveal the characteristics of the internal layer of DGNNs.
Our algorithm consists of two steps: 
(1) to handle a large number of boundaries in DGNNs, our algorithm approximates the set of critical boundaries of the query which is the given latent vector using Bernoulli dropout approach \cite{chang2018explaining};
(2) then our algorithm efficiently obtains samples which share same attributions as the query in a perspective of the trained DGNNs by expanding the tree-like exploring structure \cite{lavalle1998rapidly} until it reaches the boundaries of the region.

The advantages of our algorithm are twofold: (1) it can guarantee sample acceptance in high dimensional space where the rejection sampling based on the Monte Calro method easily fails when the region area is unknown; (2) it can handle sampling strategy in a perspective of the model where the commonly used $\epsilon$-based sampling \cite{erhan2010understanding} is not precise to obtain samples considering complex non-spherical generative boundaries \cite{laugel2019dangers}. We experimentally verify that our algorithm obtains more consistent samples compared to $\epsilon$-based sampling methods on deep convolutional GANs (DCGAN) \cite{radford2015unsupervised} and progressive growing of GANs (PGGAN) \cite{karras2017progressive}.

\section{Related Work}
\subsubsection{Generative Adversarial Networks}
The adversarial training between a generator and a discriminator has highly improved the quality and diversity of samples genereted by DGNNs \cite{goodfellow2014generative}. Many generative models have been proposed to generate room images \cite{radford2015unsupervised} and realistic human face images \cite{karras2017progressive,karras2019style}.
Despite those improvements, the generation mechanisms of the GANs are not clearly analyzed yet. 
Recent results revealed that the relationship between the input latent space and the output data space in a trained GAN by showing a manipulation in the latent vectors changes attributes in the generated data \cite{radford2015unsupervised,zhu2016generative}. 
Generation roles of some neural nodes in a trained GAN are identified with the intervention technique \cite{bau2019gan}.

\subsubsection{Explaining deep neural networks}
One can explain an output of neural networks by the sensitivity analysis, which aims to figure out which portion of an input contributes to the output. 
The sensitivity can be calculated by class activation probabilities \cite{zhou2016learning}, relevance scores \cite{montavon2017explaining} or gradients \cite{selvaraju2017grad}.
DeconvNet \cite{zeiler2014visualizing},  LIME \cite{ribeiro2016should} and SincNet \cite{ravanelli2018interpretable} trains a new model to explain the trained model.
Geometric analyis could also reveal the internal structure indirectly \cite{montufar2014number,lei2018geometric,fawzi2018empirical}.
The activation maximization \cite{erhan2010understanding}, or GANs \cite{nguyen2016synthesizing} have been used to explain the neural network by using examples.
Our method is an example-based explanation which brings a new geometric persprective to analyze DGNNs.

\subsubsection{Geometric analysis on the inside of deep neural networks}
Geometric analysis attempts to analyze the internal working process by relating the geometric properties, such as boundaries dividing the input space or manifolds along the boundaries, to the output of the model. 
The depth of a network with nonlinear activations was shown to contribute to the formation of boundary shape \cite{montufar2014number}. This property makes complex, non-convex regions surrounded by boundaries derived by internal layers. 
Although such regions are complicated, each region for a single classification in DNN classifiers is shown to be topologically connected \cite{fawzi2018empirical}. 
It has also been shown that the manifolds learned by DNNs and distributions over them are highly related to the representation capability of a network \cite{lei2018geometric}.

\subsubsection{Example-based explanation of the decision of the model}
Activation maximization is one of example-based methods to visualize the preferred inputs of neurons in a layer and according patterns in hidden layers \cite{erhan2010understanding}. The learned deep neural representation can be denoted by preferred inputs because it is related to the activation of specific neurons \cite{nguyen2016synthesizing}. The reliability of examples for explanation also has been argued considering the connectivity among the justified samples \cite{laugel2019dangers}.

\section{Generative Boundary Aware Sampling in Deep Generative Neural Networks}\label{geo_boundary}

This section presents our main contribution, the explorative generative boundary aware sampling (E-GBAS) algorithm, which can obtain samples sharing the identical attributes from the perspective of the DGNNs. Initially, we define the terms used in our algorithm. Then we explain E-GBAS which comprises of (1) an approximate representation of generative boundaries and (2) an efficient stochastic exploration to obtain samples in the complex, non-convex generative region.

\subsection{Deep Generative Neural Networks}\label{dgnn}
Although there are various architecture of DGNNs, we represent the DGNNs in a unified form.
Given DGNNs with $L$ layers, the function of DGNNs model $G$ is decomposed into $G(z)=f_{L}(f_{L-1}(\cdots (f_{1}(z)))) = f_{L:1}(z)$, where $z$ is a vector in the latent space $\mathcal{Z}\subset \mathbb{R}^{D_z}$. $f_{l:1}^i(\cdot)$ denotes the value of $i$-th element and $f_{l:1}(z) \in \mathbb{R}^{D_l}$. In general, the operation $f_l(\cdot)$ includes linear transformations and a non-linear activation function.

\subsection{Generative Boundary and Region}
The latent space of the DGNNs is divided by hypersurfaces learned during the training. The networks make the final generation based on these boundaries. We refer these boundaries as the generative boundaries.

\begin{definition}[\textbf{Generative Boundary (GB)}]
\label{def:db}
The $i$-th generative boundary at the $l$-th layer is defined as 
$$B_l^i=\{z|f_{l:1}^i(z)=0, z\in \mathcal{Z}\}.$$
\end{definition}

In general, there are numerous boundaries in the $l$-th layer of the network and the configuration of the boundaries comprises the region. Because we are mainly interested in the region created by a set of boundaries, we denote the definition of halfspace which is a basic component of the region.

\begin{definition}[\textbf{Halfspace}]
\label{def:halfspace}
Let a halfspace indicator $\mathbf{V}_l\in \{-1,0,+1\}^{D_l}$ for the $l$-th layer. 
Each element ${V_l}^i$ indicates either or both of two sides of the halfspace divided by the $i$-th GB. We define the halfspace as
\begin{gather*}
    \mathcal{H}_l^i = \begin{cases}
    \mathcal{Z} &\emph{ if }V_l^i=0\\
    \{z|{V_l}^if_{l:1}^i(z)\geq0\}&\emph{ if }{V_l}^i\in \{-1,+1\}.
    \end{cases}
\end{gather*}
\end{definition}

The region can be represented by the intersection of each halfspace in the $l$-th layer. For the case where ${V_l}^i=0$, the halfspace is defined as the entire latent space, so $i$-th GB does not contribute to comprise the region.

\begin{definition}[\textbf{Generative Region (GR)}]
Given a halfspace indicator $\mathbf{V}_l$ in the $l$-th layer, let the set of according halfspaces $\mathbf{H}=\{\mathcal{H}^1_l,\dots,\mathcal{H}_l^{D_l}\}$. Then the generative region $GR_{V_l}$ is defined as
$$GR_{V_l}=\cap_{\mathcal{H} \in \mathbf{H}}\mathcal{H}.$$
\end{definition}

For a network with a single layer ($l{=}1$), the generative boundaries are linear hyperplanes. The generative region is constructed by those boundaries and appears as a convex polytope. However, if the layers are stacked with nonlinear activation functions ($l{>}1$), the generative boundaries are bent, so the generative region will have a complicated non-convex shape \cite{montufar2014number,raghu2017expressive}.

\subsection{\textls*[-20]{Smallest Supporting Generative Boundary Set}}
Decision boundaries have an important role in classification task, as samples in the same decision region have the same class label. In the same context, we manipulate the generative boundaries and regions of the DGNNs. 

Specifically, we want to collect samples that are placed in the same generative region and have identical attributes in a perspective of the DGNNs. To define this property formally, we first define the condition under which the samples share the neural representation.

\begin{definition}[\textbf{Neural Representation Sharing (NRS)}] Given a pair of latent vectors $z_i, z_j \in \mathcal{Z}$ satisfies the neural representation sharing condition in $l$-th layer if 
$$\emph{sign}(f^k_{l:1}(z_i)) = \emph{sign}(f^k_{l:1}(z_j)), \quad \forall k\in \{1,2,\dots,D_l\}.$$
\end{definition}

\begin{figure}[h!]
    \centering
    \begin{subfigure}[t]{0.46\columnwidth}
        \includegraphics[width=\columnwidth]{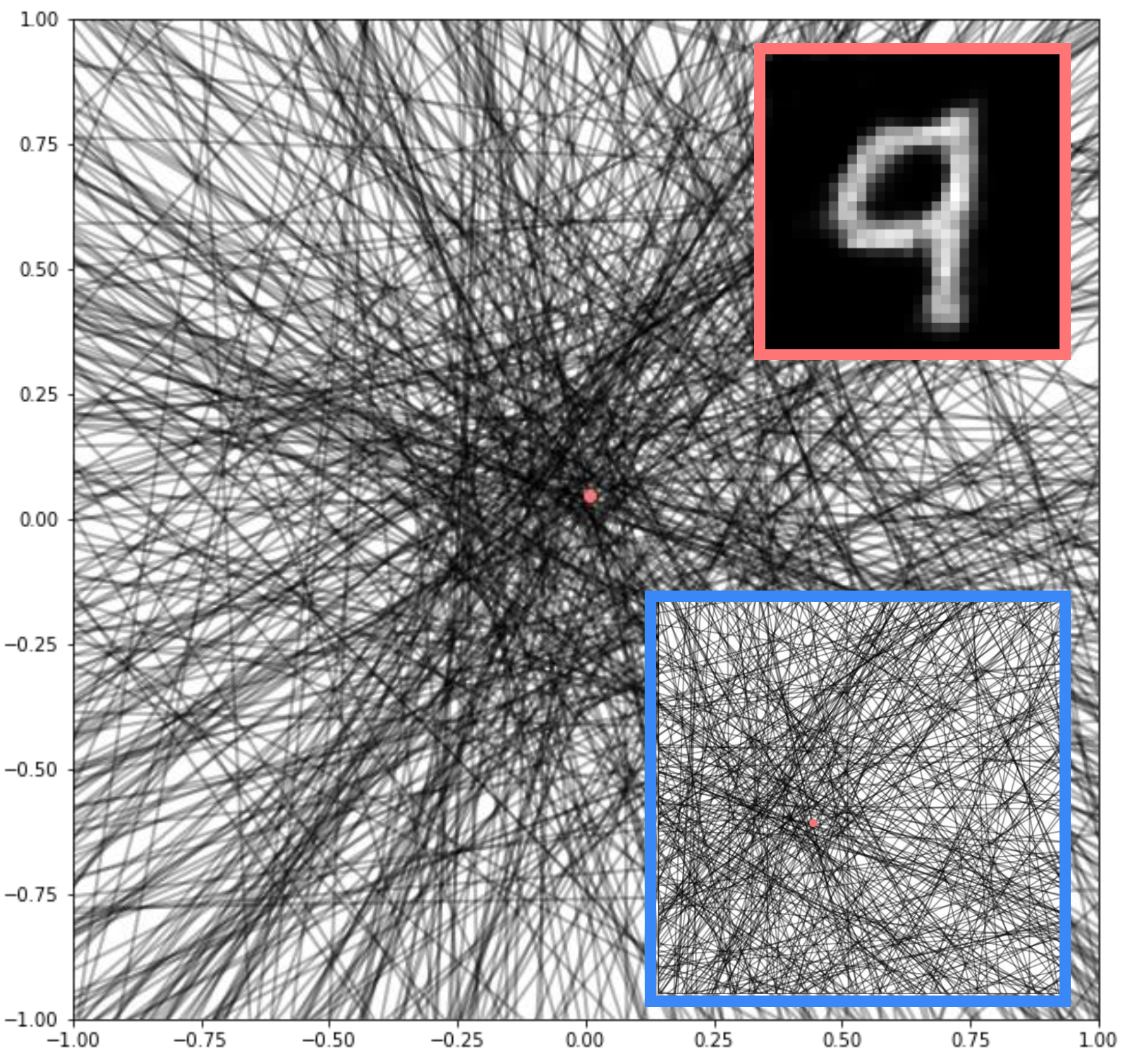}
        \caption{Original generative boundaries and generated digit image of the query}
        \label{fig:SSGBS_1}
    \end{subfigure}
    \hspace{1em}
    \begin{subfigure}[t]{0.46\columnwidth}
        \includegraphics[width=\columnwidth]{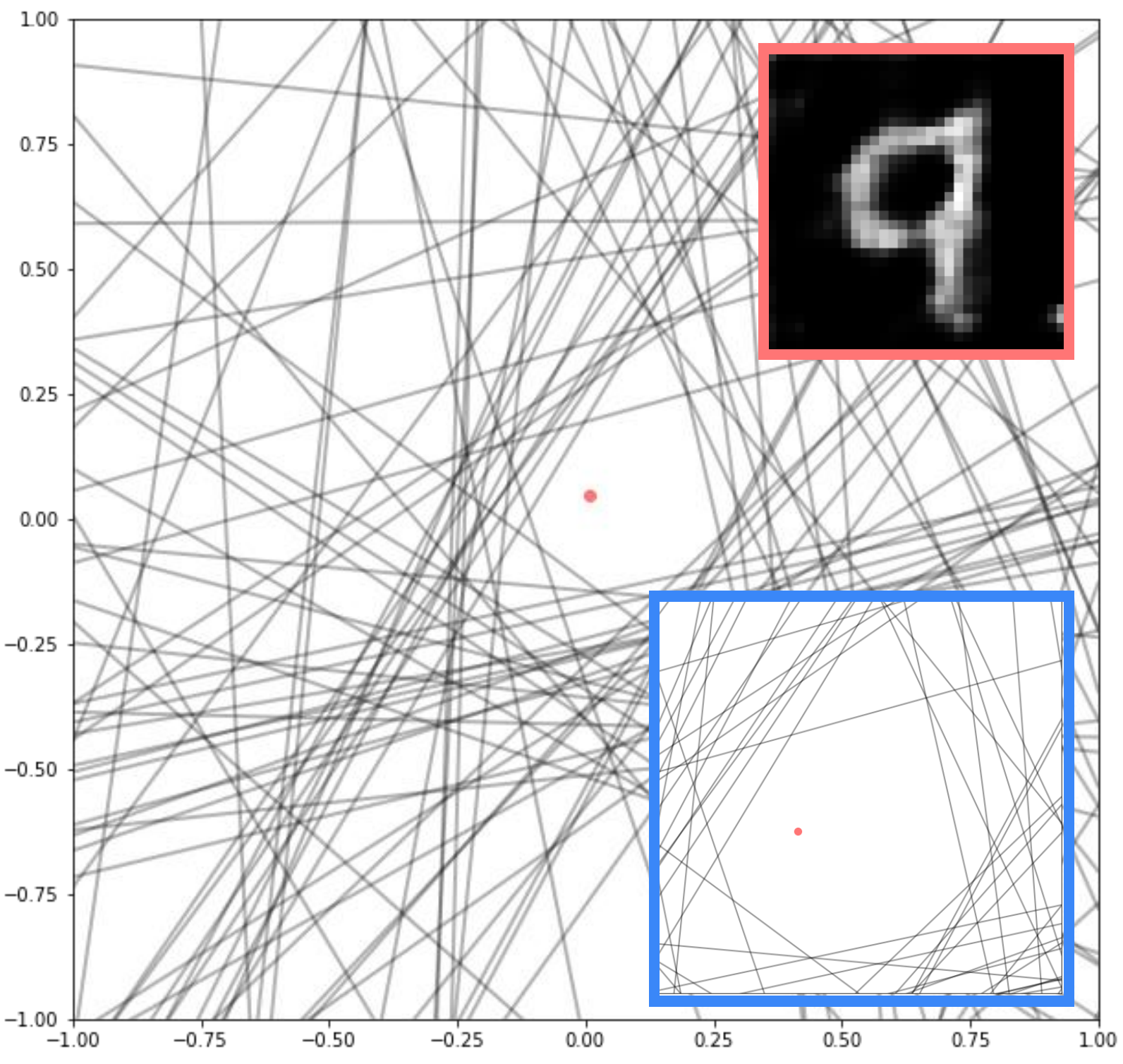}
        \caption{Generative boundaries in the SSGBS and generated digit image of the query}
        \label{fig:SSGBS_2}
    \end{subfigure}
    \caption{Results of optimization of the Bernoulli parameter $\theta$ for the given arbitrary query $z$ in the trained DCGAN on MNIST with 2-D latent space\cite{radford2015unsupervised,lecun2010mnist}. The red box denotes the generated digit image and the blue box denotes the magnified area nearby the query. (a) shows all generative boundaries in the first hidden layer ($l$=1). (b) shows SSGBS after optimization with constraint $p>0.5$.}\label{fig:SSGBS}
\end{figure}

It is practically challenging to find samples that satisfy the above condition, because a large number of generative boundaries exist in the latent space, as shown in Figure \ref{fig:SSGBS}(\subref{fig:SSGBS_1}).
Various information represented by thousands of generative boundaries makes it difficult to identify which boundary is in charge of each piece of information.
We relax the condition of the neural representation sharing by considering a set of the significant boundaries.

\begin{definition}[\textbf{\textbf{Relaxed NRS}}]
Given a subset $S$ and a pair of latent vectors $z_i, z_j \in \mathcal{Z}$ satisfies the relaxed neural representation sharing condition if $$\emph{sign}(f^k_{l:1}(z_i)) = \emph{sign}(f^k_{l:1}(z_j)), \quad \forall k\in S\subset \{1,2,\dots,D_l\}.$$
\end{definition}

Then, we must select important boundaries for the relaxed NRS in the $l$-th layer. We believe that not all nodes deliver important information for the final output of the model as some nodes could have low relevance of information \cite{morcos2018importance}. Furthermore, it has been shown that a subset of features mainly contributes to the construction of outputs in GAN \cite{bau2019gan}. We define the \textit{smallest supporting generative boundary set (SSGBS)}, which minimizes the influences of minor (non-critical) generative boundaries.

\begin{definition}[\textbf{\textls*[-40]{Smallest Supporting Generative Boundary Set}}]

Given the generator $G$ and a query $z_0\in \mathcal{Z}$, for $l$-th layer and any real value $\delta>0$, 
if there exists an indicator $\mathbf{V}_l^*$ such that
\begin{align*}
    \left\lVert G(z)-G(z_0)\right\rVert\leq\delta, \quad z\in \{z|f_{l-1:1}(z)\in GR_{\mathbf{V}_l^*}\}
\end{align*}
and there is no $\mathbf{V}_l'$ where $\lVert\mathbf{V}_l'\rVert_1<\lVert\mathbf{V}_l^*\rVert_1$ such that 
\begin{align*}
    \left\lVert G(z')-G(z_0)\right\rVert\leq\delta, \quad z'\in \{z'|f_{l-1:1}(z')\in GR_{\mathbf{V}_l'}\}
\end{align*}
then we denote a set $\mathcal{B}_{\mathbf{V}_l^*}=\{B_l^i|{V^*_l}^i\neq0, i\in\{1,2,\dots,D_l\}\}$ as the smallest supporting generative boundary set (SSGBS). 

In the same context, we denote the generative region $GR_{\mathbf{V}_l^*}$ that corresponds to the SSGBS as the smallest supporting generative region (SSGR).
\label{def:ssdbs}
\end{definition}

It is impractical to explore all the combinations of boundaries to determine the optimal SSGBS, owing to the exponential combinatoric search space.\footnote{For example, a simple fully connected layer with $N$ outputs generates up to $2^N$ generative boundary sets.} To avoid this problem, we used the Bernoulli dropout approach \cite{chang2018explaining} to obtain the SSGBS. We define this dropout function as $\phi(h,\theta)=h{\odot} m,\ m{\sim} Ber(\theta)$, where $\odot$ is an element-wise multiplication. We optimize $\theta$ to minimize the loss function $L$, which quantifies the degradation of generated image with the sparsity of Bernoulli mask.

\begin{align}
    \theta^*&=\arg\min_{\theta} L(z_0, l, \theta)\\
    &=\arg\min_{\theta} \left\lVert f_{L:l+1}(\phi(f_{l:1}(z_0),\theta)){-}G(z_0)\right\rVert{+}\lambda\lVert \theta\rVert_1\nonumber
    \label{eq:loss}
\end{align}

We iteratively update the parameter using gradient descent to minimize the loss function in Equation (\ref{eq:loss}). Then we obtain the SSGBS $\mathcal{B}_{\mathcal{V}_l^*}$ from the optimized Bernoulli parameter $\theta$ with a proper threshold in the $l$-th layer. For each iteration, we apply the element-wise multiplication between $f_{l:1}(z_0)$ and sampled mask $m{\sim}Ber(\theta)$ to obtain masked feature value and feed it to obtain the modified output.

\begin{algorithm}[h]
\caption{Bernoulli Mask Optimization (BerOpt)}\label{alg:BerMaskOpt}
Input: $z_0$: a query, $G(.)=f_{L:1}(.)$: a DGNN model, \\$l$: a target layer\\
Output: $\theta$: the optimized Bernoulli parameter for SSGBS

\begin{algorithmic}[1]
\STATE Initialize $\theta\in[0,1]^{D_l}$ 
\STATE $h_0=f_{l:1}(z_0)$
\WHILE{not converge}
  \STATE Sample $m{\sim}Ber(\theta)$
  \STATE $h_m=h_0\odot m$
  \STATE $x_0=f_{L:l+1}(h_0)$, $x_m=f_{L:l+1}(h_m)$
  \STATE Compute loss $L(z_0, l, \theta)$
  \STATE Update $\theta$ with $\nabla_\theta L$
\ENDWHILE
\STATE \textbf{return} $\theta$
\end{algorithmic}
\end{algorithm}

After obtaining the optimal Bernoulli parameter $\theta^*$, we first define an optimal halfspace indicator $\mathbf{V}_l^*$ with the proper probability threshold (e.g., $p=0.5$). We set the value of elements in $\mathbf{V}_l^*$ to zero for removing GBs which have minor contributions to the generation mechanism. That is, 
\begin{align*}
    {V_l^*}^i=\mathbb{I}(\theta^*>p) \cdot \text{sign}(f_{l:1}^i(z_0))
\end{align*}
where $\mathbb{I}$ is indicator function. Representing SSGBS $\mathcal{B}_{\mathcal{V}_l^*}$ and SSGR $GR_{\mathcal{V}_l^*}$ is straightforward from the Definition \ref{def:ssdbs} with $\mathbf{V}_l^*$. Figure \ref{fig:SSGBS} shows the generative boundaries and the generated digit of the SSGBS without and with the optimized Bernoulli parameter $\theta^*$ with $p>0.5$.
The generated digits indicate that the effect of the removal of minor generative boundaries on the output is not significant.

\subsection{\textbf{\textls*[-55]{Explorative Generative Boundary Aware Sampling}}}

\begin{figure}[h!]
    \centering
    \begin{subfigure}[t]{0.47\columnwidth}
    \includegraphics[width =\textwidth]{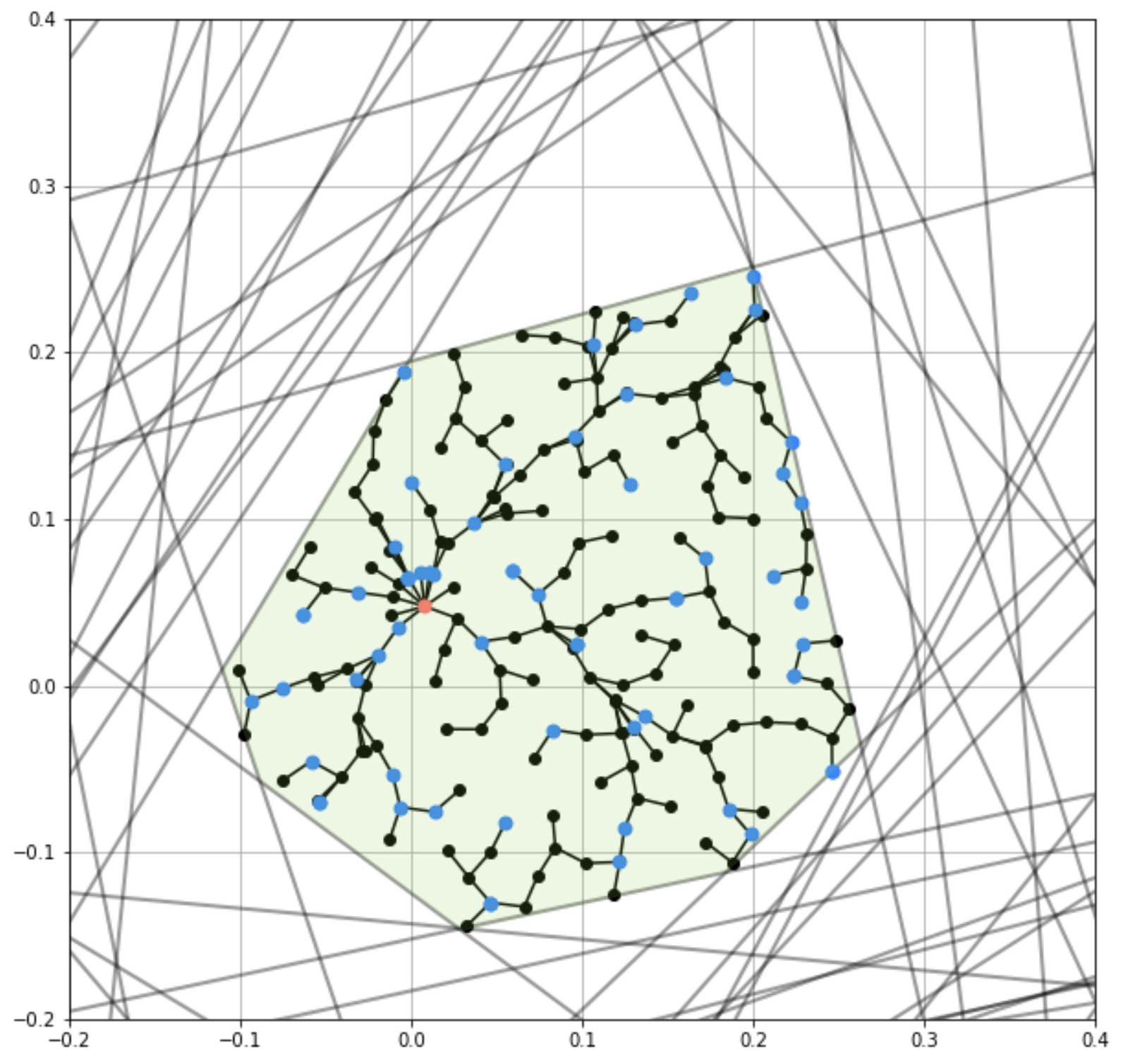}
    \caption{}
    \label{fig:RRT1_a}
    \end{subfigure}
    \begin{subfigure}[t]{0.44\columnwidth}
    \includegraphics[width =\textwidth]{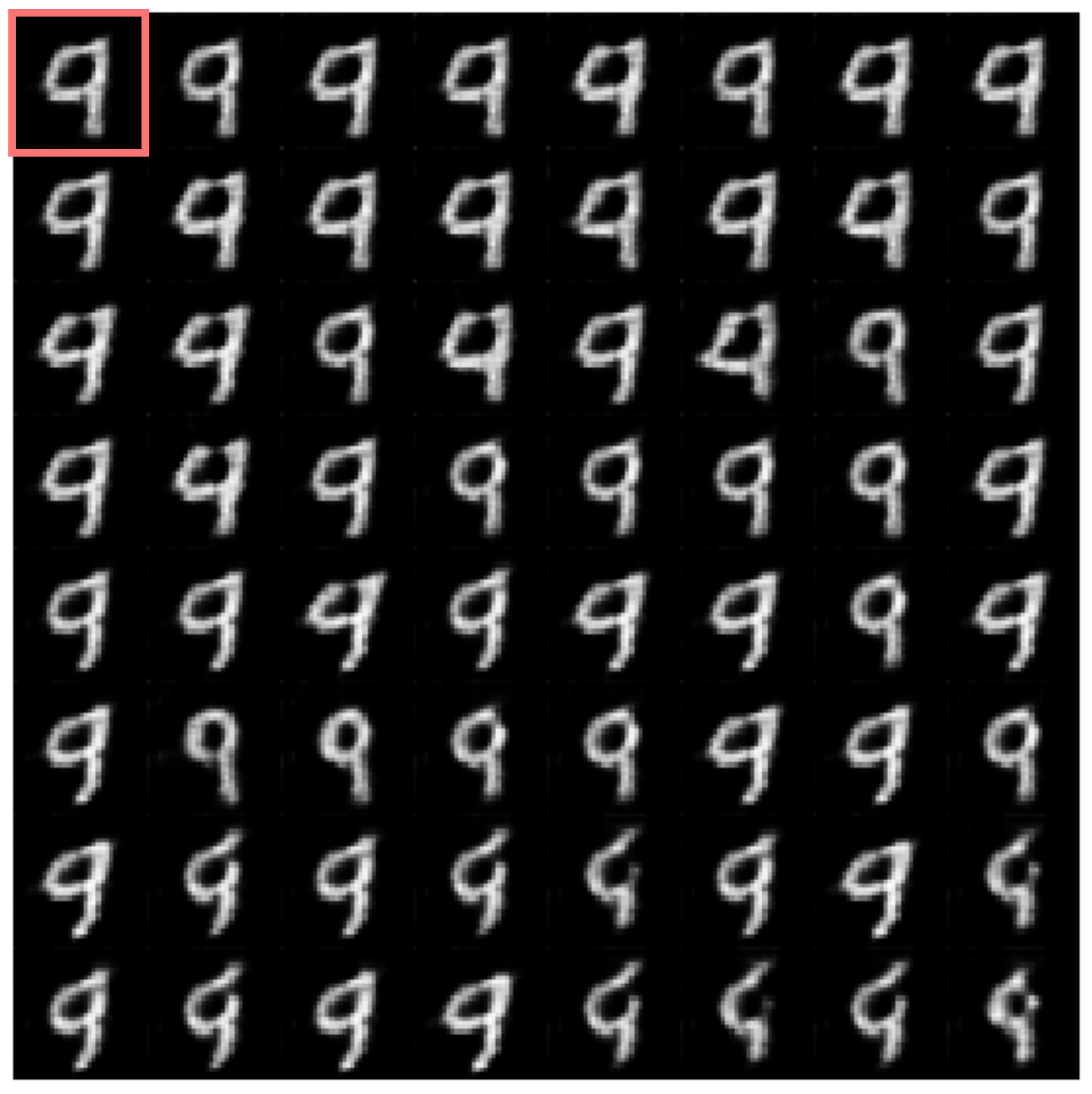}
    \caption{}
    \label{fig:RRT1_b}
    \end{subfigure}
    \caption{(a) Visualization of explorative trajectories of GB-RRT for a given query (red dot) in the first hidden layer ($l{=}1$) of DCGAN-MNIST and (b) generated outputs from uniform randomly chosen samples (blue dot in (a)). The red box denotes the generated output of the query.
    }
    \label{fig:RRT}
\end{figure}

After obtaining SSGR $GR_{\mathcal{V}_l^*}$, we gather samples in the region and compare the generated outputs of them. Because the $GR_{\mathcal{V}_l^*}$ possesses a complicated shape, simple neighborhood sampling methods such as $\epsilon$-based sampling cannot guarantee exploration inside the $GR_{\mathcal{V}_l^*}$. To guarantee the relaxed NRS, we apply the GB constrained exploration algorithm inspired by the rapidly-exploring random tree (RRT) algorithm \cite{lavalle1998rapidly}, which is invented for the robot path planning in complex configuration spaces. We refer to the modified exploration algorithm as generative boundary constrained RRT (GB-RRT). Figure \ref{fig:RRT}(\subref{fig:RRT1_a}) depicts the explorative trajectories of GB-RRT.

\begin{algorithm}[h]
\caption{Generative boundary constrained rapidly-exploring random tree (GB-RRT)}\label{alg:RRT}
Input: $z_0$: a query, $GR_{\mathbf{V}_l^*}$: SSGR\\
Parameters: $I$: a sampling interval, \\
$N$: a maximum number of iterations, $\delta$: a step size\\
Output: $Q$: samples in the SSGR

\begin{algorithmic}[1]
\STATE Initialize queue $Q_0=\{z_0\}$
\FOR{$i=1 \dots N$}
  \STATE Sample $z_i\sim U(z_0-I,z_0+I)$
  \STATE $q_i = nearest(Q_{i-1}$,$z_i$)
  \STATE $z_i'=(z_i-q_i)/\lVert z_i-q_i\rVert * \delta + q_i$
  \STATE \algorithmicif{ $z'\in GR_{\mathbf{V}_l^*}$ \textbf{and} $\left\lVert z'- nearest(Q_{i-1},z') \right\rVert {>} \delta$}
  \STATE \algorithmicthen{ $Q_i=Q_{i-1}\cup\{z'\}$}
\ENDFOR
\STATE \textbf{return} $Q_N$
\end{algorithmic}
\end{algorithm}

We name the entire sample exploration process for DGNNs which is comprised of finding SSGBS in arbitrary layer and efficiently sampling in SSGR as explorative generative boundary aware sampling (E-GBAS).

\begin{algorithm}[h!]
\caption{Explorative generative boundary aware sampling (E-GBAS)}\label{alg:metaAlgo}
Input: $z_0$: a query, $G(.)=f_{L:1}(.)$: DGNN model,
\\$l$: a target layer, $p$: threshold for SSGBS selection\\
Output: $Z$: a set of samples in the same SSGR of $z_0$

\begin{algorithmic}[1]
\STATE Optimize $\theta^*=\textbf{BerOpt}(z_0, G, l)$
\STATE Compute $\mathbf{V}_l^*=[{V_l^*}^1,\dots$ ${V_l^*}^{D_l}]^T$\\
\quad\quad\quad\quad where ${V_l^*}^i=\mathbb{I}(\theta^*>p) \cdot sign(f_{l:1}(z_0))$
\STATE Derive $GR_{\mathbf{V}_l^*}$
\STATE Sample a set $Z=\textbf{GB-RRT}(z_0, GR_{\mathbf{V}_l^*})$
\STATE \textbf{return} $Z$
\end{algorithmic}
\end{algorithm}

\section{Experimental Evaluations}
This section presents analytical results of our algorithm and empirical comparisons with variants of $\epsilon$-based sampling method. We select three different DGNNs; (1) DCGAN \cite{radford2015unsupervised} with the wasserstein distance \cite{arjovsky2017wasserstein} trained on MNIST, (2) PGGAN \cite{karras2017progressive} trained on the church dataset of LSUN \cite{yu2015lsun} and (3) the celebA dataset \cite{liu2015faceattributes}.

\begin{figure}[!h]
    \centering
    \begin{subfigure}[t]{0.45\columnwidth}
    \includegraphics[width =\textwidth]{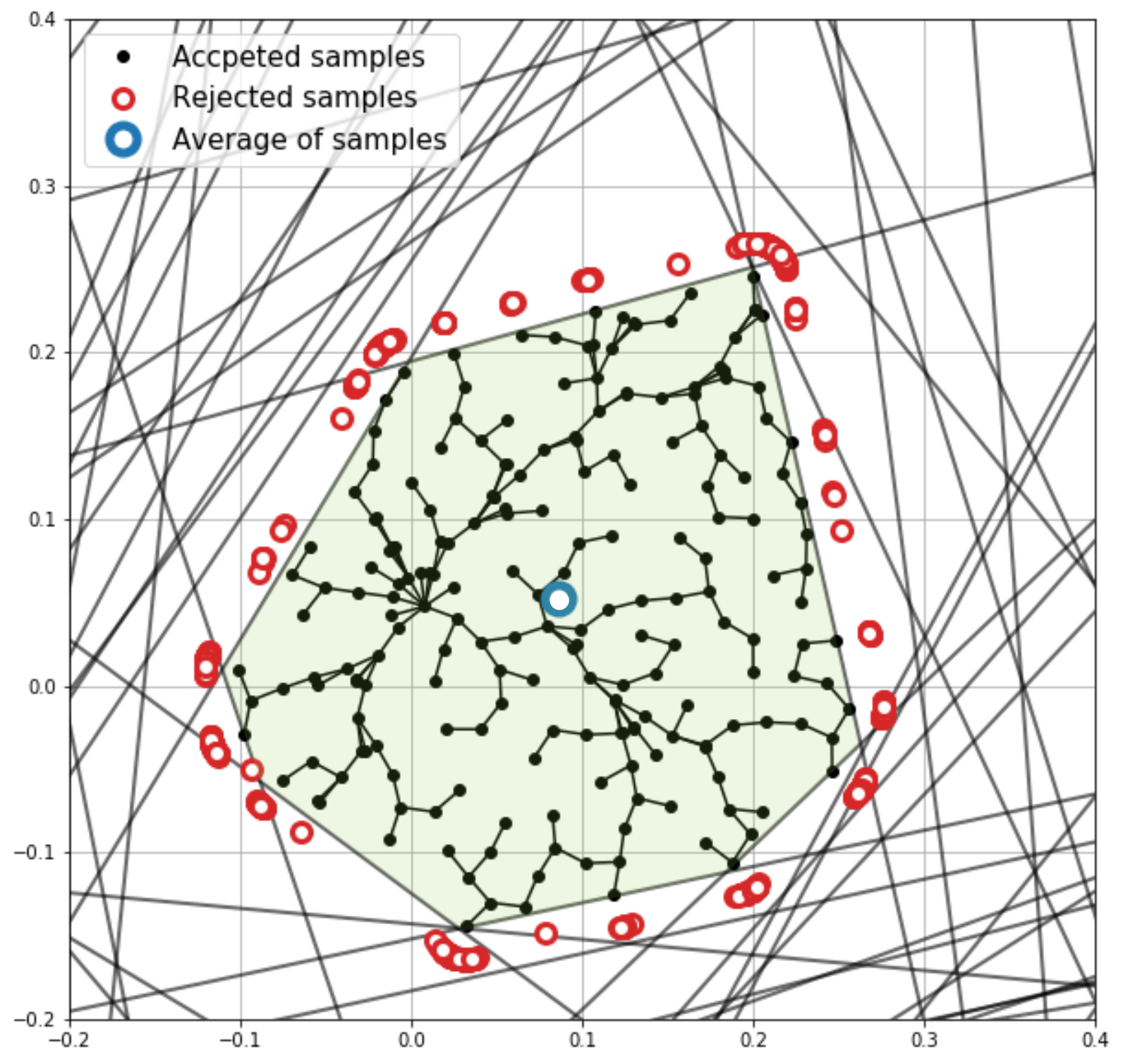}
    \caption{}
    \label{fig:seteps1}
    \end{subfigure}
    \begin{subfigure}[t]{0.45\columnwidth}
    \includegraphics[width =\textwidth]{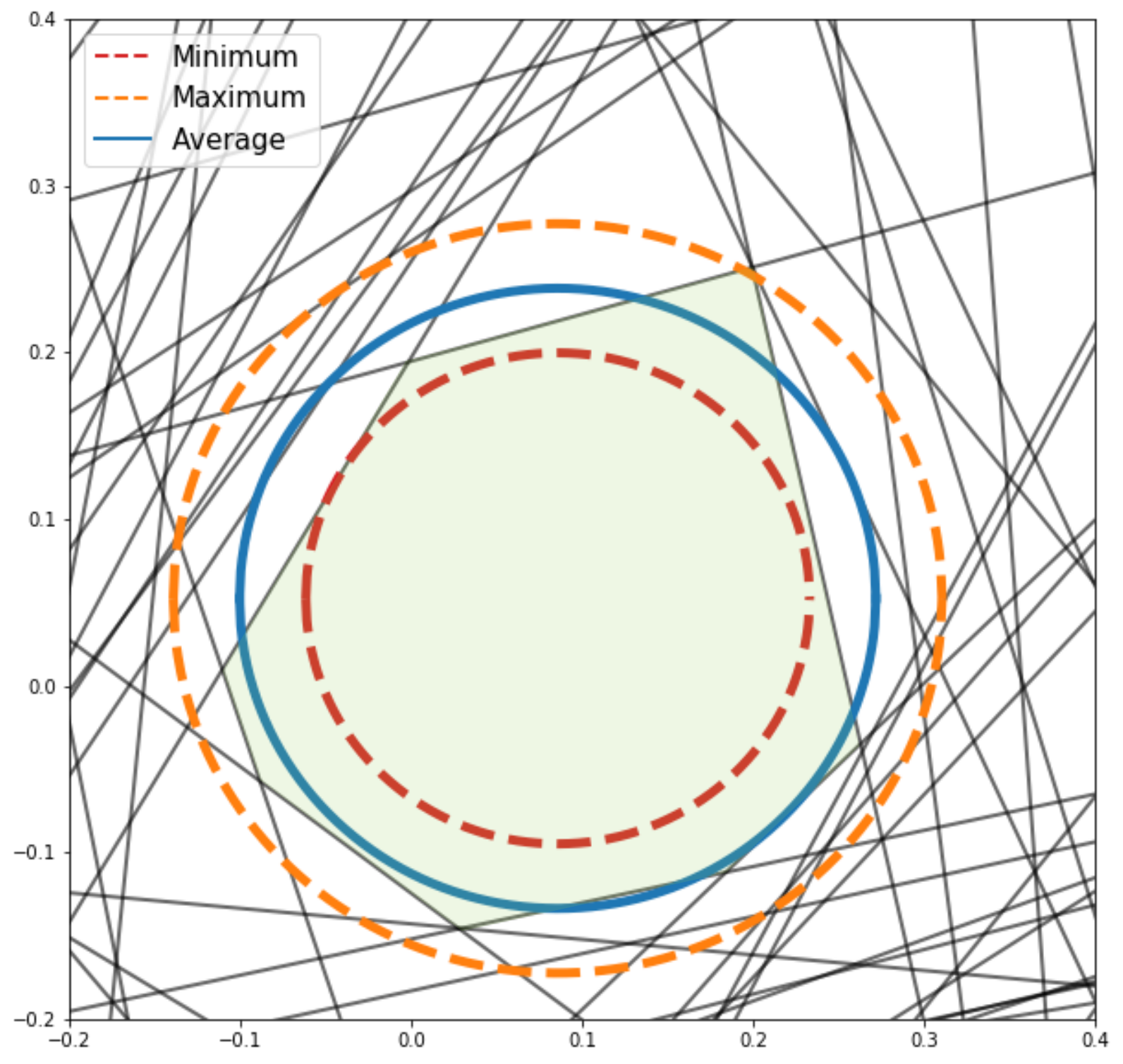}
    \caption{}
    \label{fig:seteps2}
    \end{subfigure}
    \caption{(a) The accepted samples (black dots), rejected samples (red dots) and average of the accepted samples (blue dot) by E-GBAS. (b) Visualizing the selection of $\epsilon_{L_2}$ to make the area close to that of SSGR. 
    The $\epsilon_{L_2}$-balls of each distance. min (red), max (orange) and average of min/max distance (blue).}
    \label{fig:SetEps}
\end{figure}

\begin{figure}[!h]
    \centering
    \begin{subfigure}[t]{0.32\columnwidth}
    \includegraphics[width =\textwidth]{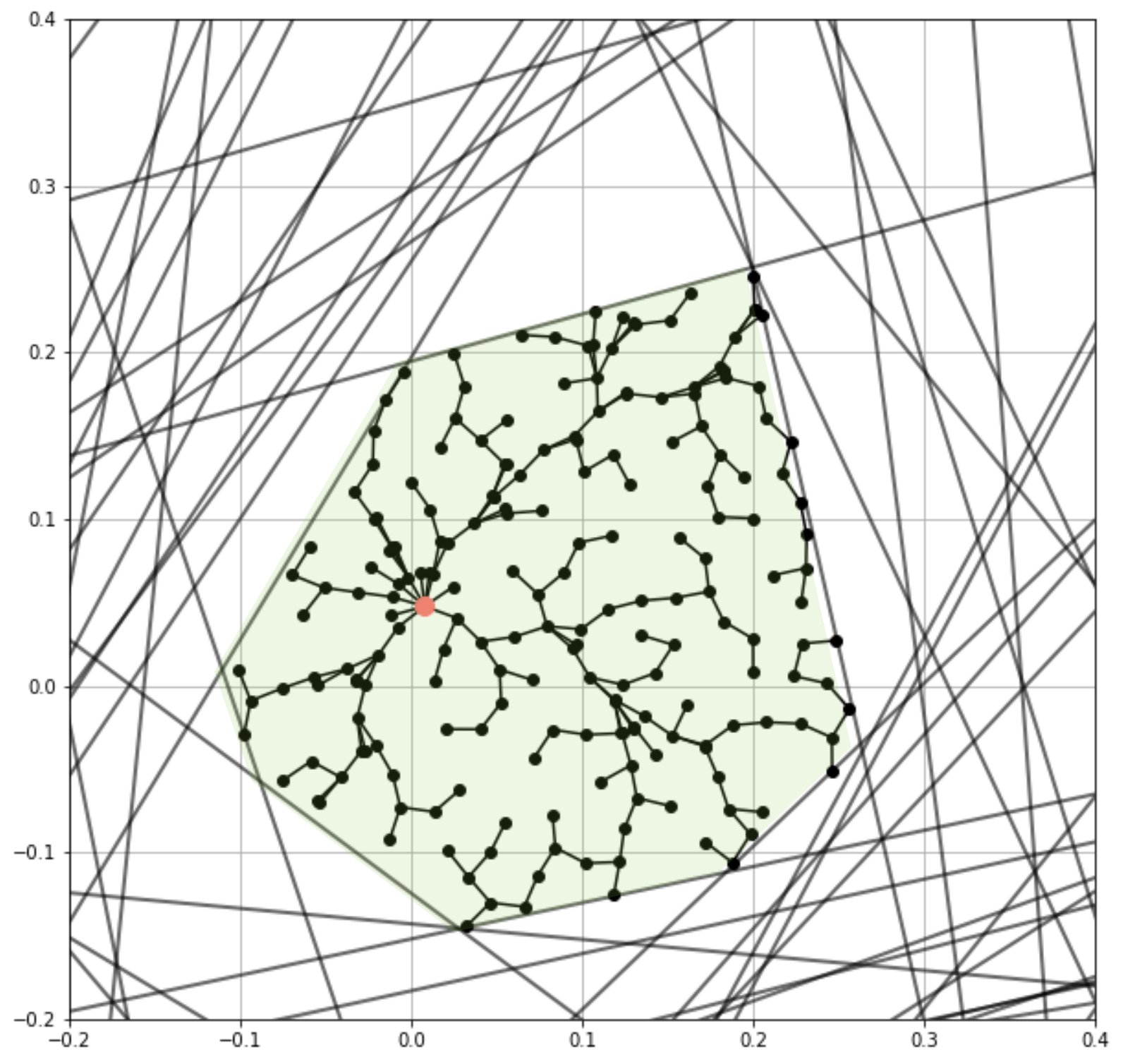}
    \caption{E-GBAS}
    \label{fig:eps0}
    \end{subfigure}
    \begin{subfigure}[t]{0.32\columnwidth}
    \includegraphics[width =\textwidth]{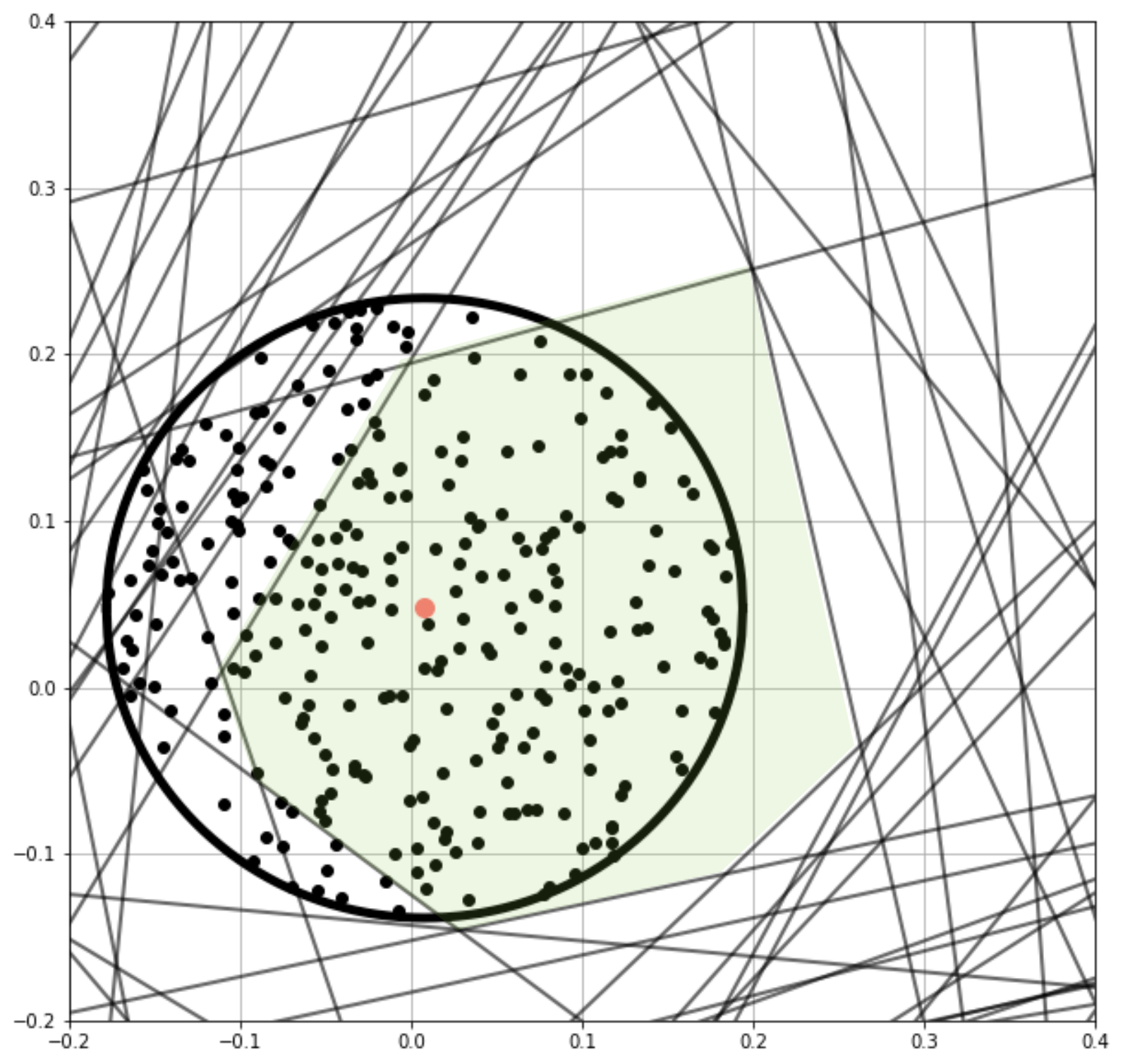}
    \caption{$\epsilon_{L_2}$}
    \label{fig:eps2}
    \end{subfigure}
    \begin{subfigure}[t]{0.32\columnwidth}
    \includegraphics[width =\textwidth]{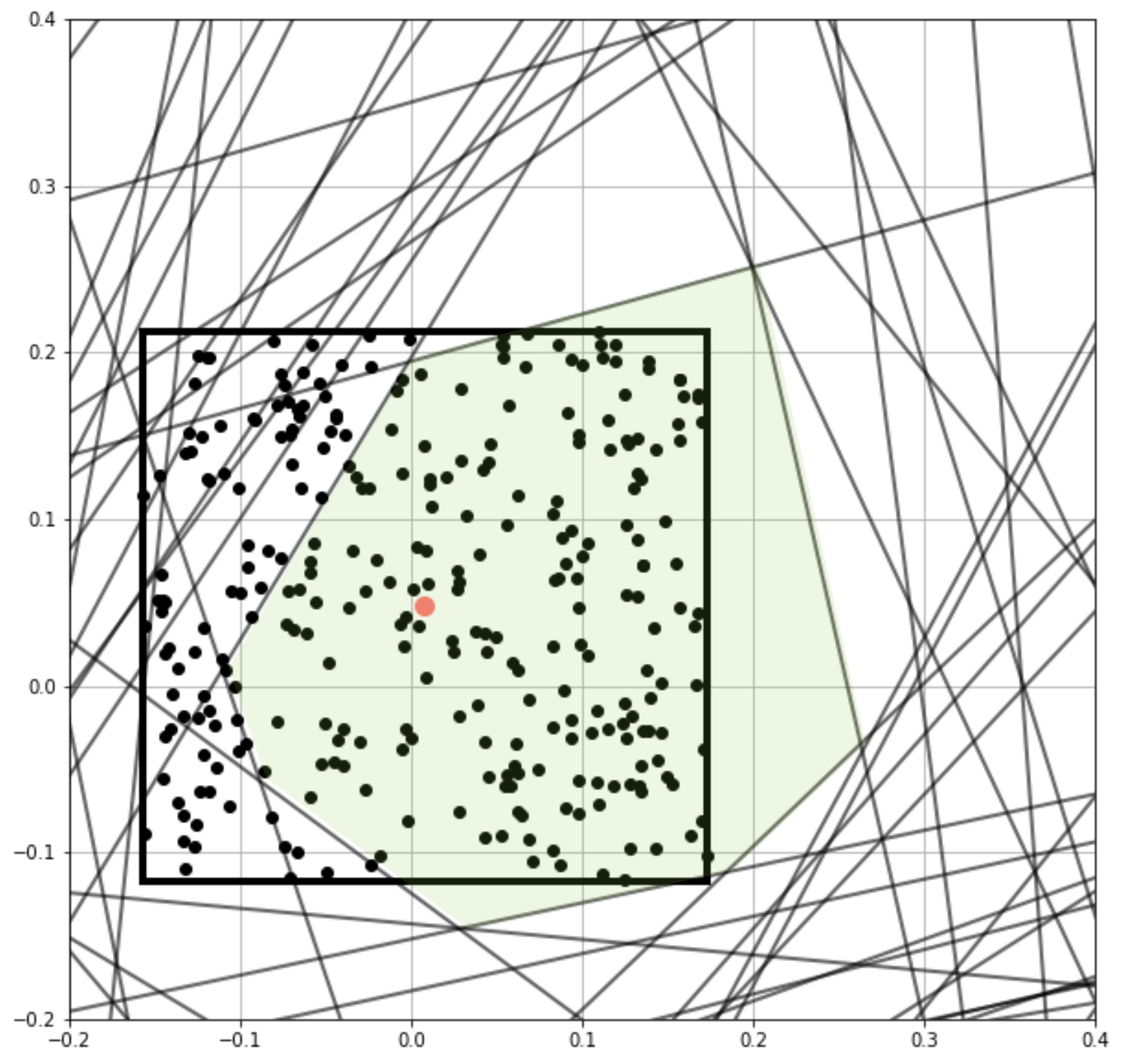}
    \caption{$\epsilon_{L_\infty}$}
    \label{fig:eps3}
    \end{subfigure}
    
    \caption{Geometric comparison of (a) E-GBAS, (b) $\epsilon_{L_2}$ and (c) $\epsilon_{L_\infty}$-based sampling methods. 
    Although the two $\epsilon$-balls cover some area of the GR, they cannot cover all of the GR and have a possibility to include infeasible area. Whereas, the E-GBAS includes the only feasible area of GR for sampling.}
    \label{fig:SamplingMethods}
\end{figure}

The $\epsilon$-based sampling collects samples based on $L_p$ distance metric. We choose $L_2$ and $L_\infty$ distance as baseline, and sample in each $\epsilon$-ball centered at the query.
use
In practice, the value of $\epsilon$ is manually selected. We use the set of accepted samples and rejected samples, $Z_{accept}$ and $Z_{reject}$, obtained by the E-GBAS to set the $\epsilon$ for fair comparisons. 
We set the average of accepted samples $z_{avg}$ which can represent the middle point of the SSGR, then we calculate $\epsilon_{L_2}$ with min/max distance between $z_{avg}$ and $Z_{reject}$ as, 
\begin{gather*}
    \epsilon_{L_2} = \frac{1}{2}\left(\max_{z\in Z_{reject}}\lVert z_{avg}-z\rVert + \min_{z\in Z_{reject}}\lVert z_{avg}-z\rVert\right).
\end{gather*}
Figure \ref{fig:SetEps} indicates the visualization of calculating $\epsilon$ in the DCGAN-MNIST. After $\epsilon_{L_2}$ is set, $\epsilon_{L_\infty}$ are determined to have the same volume as the $\epsilon_{L_2}$-ball. Figure \ref{fig:SamplingMethods} shows the geometric comparisons of each sampling method in the first hidden layer ($l{=}1$) of DCGAN-MNIST.

\subsection{Qualitative Comparison of E-GBAS and $\epsilon$-based Sampling}
\begin{figure*}[h!]
    \centering
        \includegraphics[width=\textwidth]{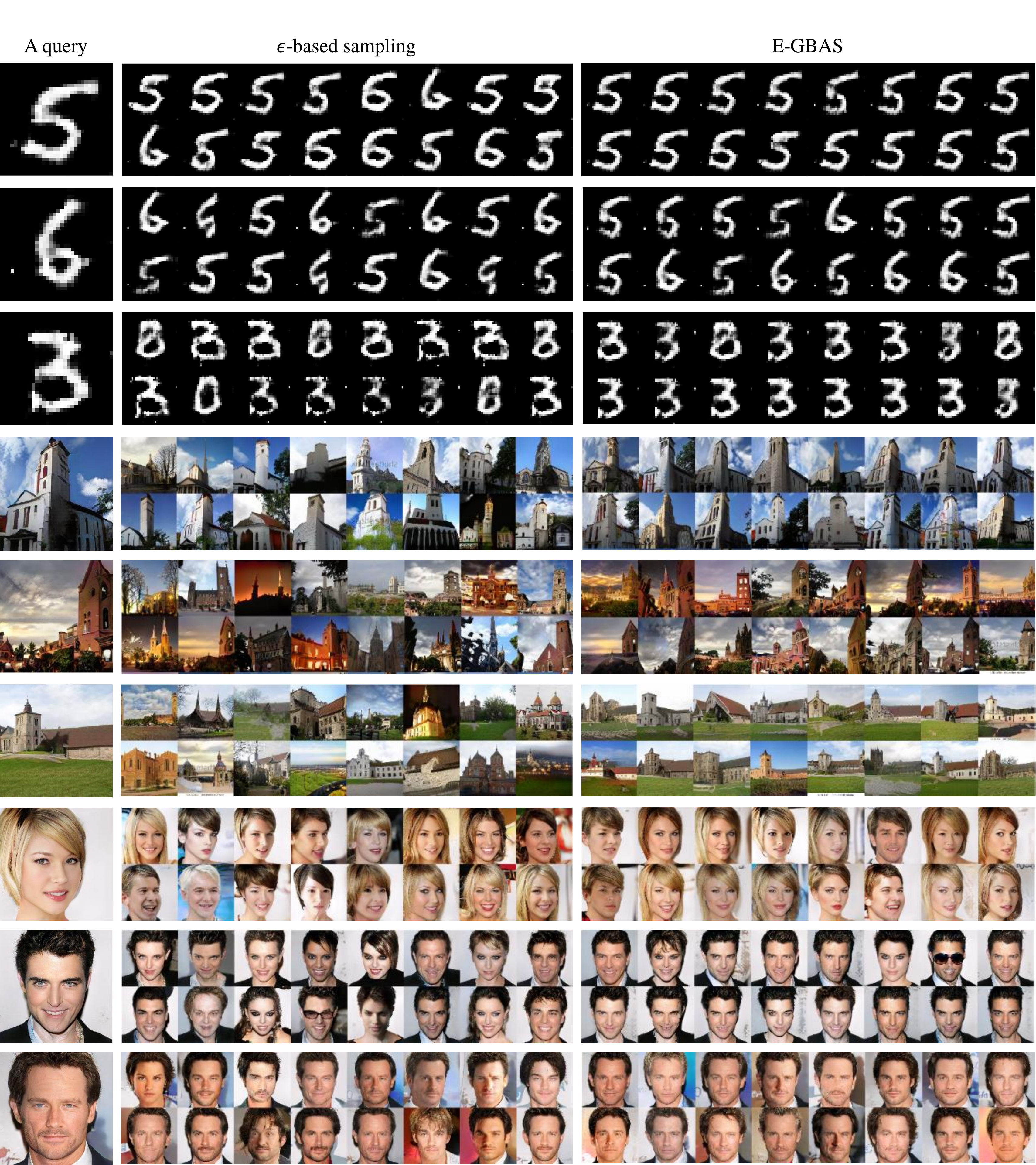}
        \label{fig:celebA}
    \caption{Generated samples from a query input (left), $\epsilon$-based sampling (middle) and E-GBAS sampling (right) of the three DGNNs (DCGAN-MNIST, PGGAN-LSUN and PGGAN-celebA.). We confirm that the generated images by E-GBAS have more consistent information compared to the $\epsilon_{L_2}$-based sampling.}\label{fig:gen_fig}
\end{figure*}

We first demonstrate how the generated samples vary if they are inside or outside of the obtained GR. As shown in Figure \ref{fig:main_fig}, we mainly compare the samples generated from E-GBAS (blue region) to the samples from the $\epsilon_{L_2}$-based sampling (red region).
A given query and a target layer, E-GBAS explores the SSGR and obtains samples that satisfy the relaxed NRS. Figure \ref{fig:gen_fig} depicts the results of the generated images from E-GBAS and the $\epsilon_{L_2}$-based sampling. We observed that the images generated by E-GBAS share more consistent attributes (e.g., composition of view and hair color) which is expected property of NRS.
For example, in the first row of celebA results, we can identify the sampled images share the hair color and angle of face with different characteristics such as hair style. In LSUN dataset, the second row of results share the composition of buildings (right aligned) and the weather (cloudy).

\begin{figure}[h!]
    \centering
    \begin{subfigure}[t]{1\columnwidth}
    \includegraphics[width =\textwidth]{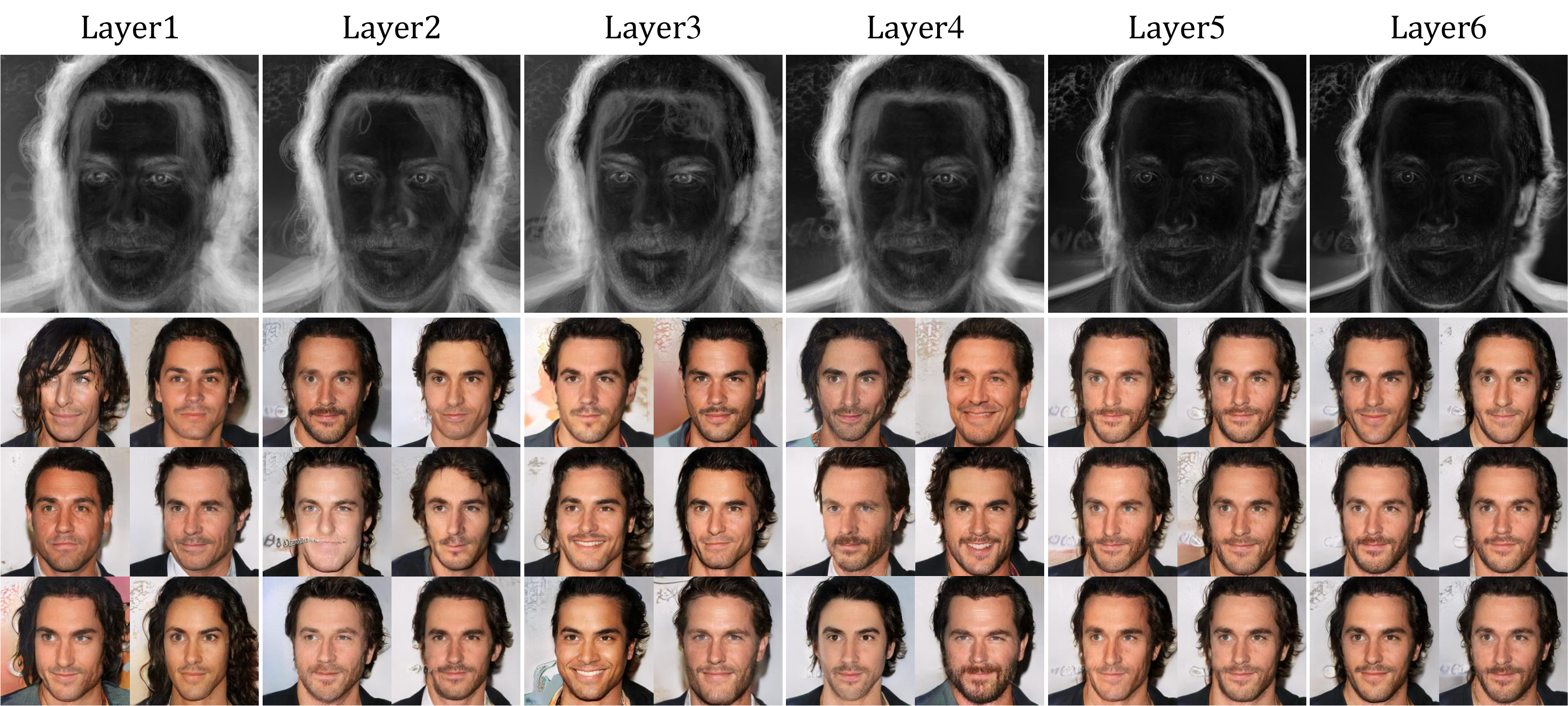}
    \caption{PGGAN-LSUN}
    \end{subfigure}
    
    \begin{subfigure}[t]{1\columnwidth}
    \includegraphics[width =\textwidth]{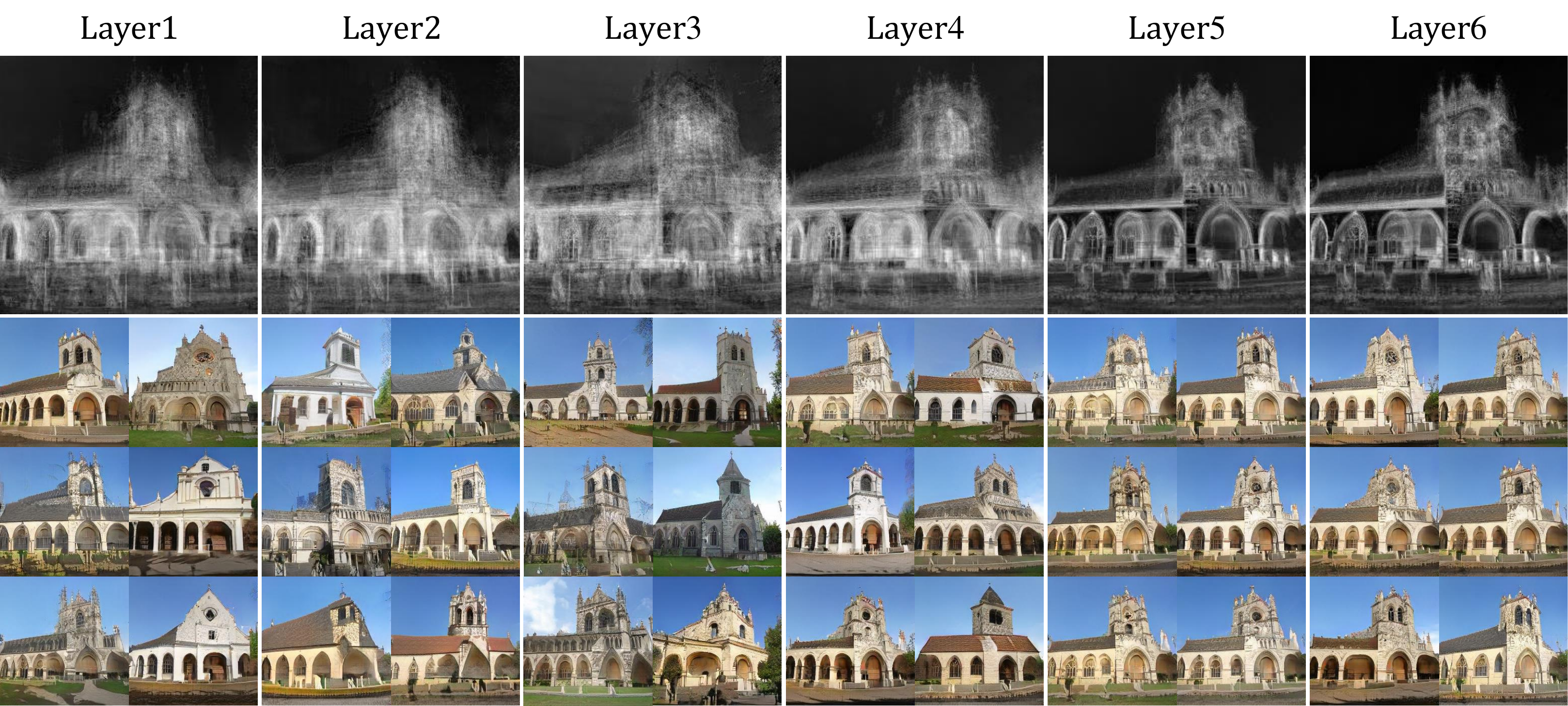}
    \caption{PGGAN-celebA}
    \end{subfigure}
    \caption{Examples of variations of generated images for each target layer. The first row shows the standard deviations of generated images for each target layer.}
    \label{fig:layer_depth}
\end{figure}

We try to analyze the generative mechanism of DGNNs along the depth of layer by changing the target layer.
Figure \ref{fig:layer_depth} shows the examples and the standard deviations of the generated images by E-GBAS in each target layer.
From the results, we discover that the variation of images is more localized as the target layer is set to be deeper. We argue that the GB in the lower layer attempts to maintain an abstract and generic information (e.g., angle of scene/entire shape of face), while those in the deeper layer tends to retain a concrete and localized information (e.g., edge of wall/mustache).

\begin{table*}[!t]
\centering
\begin{tabular}{c||c|c|c|c||c|c|c||c|c|c}

\hline\hline
\multicolumn{1}{l||}{} & \multicolumn{4}{c||}{DCGAN-MNIST} & \multicolumn{3}{c||}{PGGAN-LSUN}        & \multicolumn{3}{c}{PGGAN-celebA} \\ \hline
Layer \#           & 1      & 2      & 3      & 4     & 2           & 4     &6        & 2               & 4 &6               \\ \hline\hline
$\epsilon_{L_2}$-based sampling       & 0.0819   & 0.0711   & 0.0718   & 0.0343  & 0.4951      & 0.4971   & 0.4735     & 0.5150          & 0.4994   &0.4892       \\ \hline
$\epsilon_{L_\infty}$-based sampling       & 0.0834   & 0.0722   & 0.0720   & 0.0344  & 0.4641      & 0.4322  &0.3365      & 0.4859          & 0.4799   &0.3384       \\ \hline
E-GBAS                 & \textbf{0.0781}   & \textbf{0.0694}   & \textbf{0.0675}   & \textbf{0.0323}  & \textbf{0.3116}      & \textbf{0.2558}   & \textbf{0.1748}     & \textbf{0.2980}          & \textbf{0.2789}   & \textbf{0.1446}       \\ \hline\hline
\end{tabular}
\caption{Standard deviations of generated images in each sampling methods. The number indicates the index of layer that GB constraint is applied in each DGNNs, where higher number is close to the output generation. The E-GBAS shows the lowest standard deviations compared to $\epsilon$-based sampling methods.}\label{tb:exp1}
\end{table*}

\subsection{Quantitative Results}
\subsubsection{The Similarity of Activation Values in Discriminator}
A DGNN with the adversarial training has a discriminator to measure how realistic is the output created from a generator. During the training, the discriminator learns features to judge the quality of generated images. In this perspective, we expect that generated outputs from samples which satisfy NRS have similar feature values in the internal layers of the discriminator.
We use cosine similarity between feature values of samples and the query. The relative evaluations of NRS for each sampling method are calculated by the average of similarities. When we denote a discriminator $D(X)=d_{L:1}(X)$, the query $z_0$ and the obtained set of samples $Z$, the similarity of feature values in the $l$-th layer is defined as the Equation (\ref{eq:metricDis}). The operation $d$ consists of linear transformations and a non-linear activation function.
\begin{equation}\label{eq:metricDis}
    \textit{S}\textsubscript{\textit{$d_l$}} = E_{z\in Z}\Bigg[ \frac{d_l(G(z))^T d_l(G(z_0))}{\left\lVert d_l(G(z))\right\rVert  \left\lVert d_l(G(z_0))\right\rVert}\Bigg], \ l\in\{1,2,\dots,L\}
\end{equation}
Table 2 shows the results of measuring the similarity for each internal layer in the discriminator.

\begin{table}[!h]
\centering
\begin{tabular}{c|c||c|c|c|c}
\hline\hline
&Layer \#&  1  &  2 &  3 &  4\\ \hline\hline
\multirow{3}{*}{\rot{90}{\small{MNIST}}}&$\epsilon_{L_2}$-based& 0.722   & 0.819   & 0.864   & 0.991  \\ \cline{2-6}
&$\epsilon_{L_\infty}$-based      & 0.727   & 0.823   & 0.867   & 0.991  \\ \cline{2-6}
&E-GBAS                 & \textbf{0.747}   & \textbf{0.838}   & \textbf{0.878}   & \textbf{0.992}  \\ \hline\hline

\multirow{3}{*}{\rot{90}{\small{LSUN}}}&$\epsilon_{L_2}$-based       & 0.578   & 0.602   & 0.957   & 0.920  \\ \cline{2-6}
&$\epsilon_{L_\infty}$-based       & 0.551   & 0.613   & 0.960   & 0.946  \\ \cline{2-6}
&E-GBAS                 & \textbf{0.578}   & \textbf{0.637}   & \textbf{0.967}   & \textbf{1.000}  \\ \hline\hline

\multirow{3}{*}{\rot{90}{\small{celebA}}}&$\epsilon_{L_2}$-based       & 0.678   & 0.718   & 0.785   & 0.963  \\ \cline{2-6}
&$\epsilon_{L_\infty}$-based       & 0.684   & 0.720   & 0.789   & 0.965  \\ \cline{2-6}
&E-GBAS                 & \textbf{0.702}   & \textbf{0.733}   & \textbf{0.804}   & \textbf{0.970}  \\ \hline\hline
\end{tabular}
\caption{Comparisons of the average cosine similarity of feature values of the discriminator. The number indicates the index of layer in a discriminator.}
\end{table}\label{tb:dis}
\subsubsection{Variations of Generated Image}
To quantify the consistency in attributes of the generated images, we calculate the standard deviation of generated images sampled by E-GBAS and variants of the $\epsilon$-based sampling.
The standard deviation is calculated as Equation (\ref{eq:cmpMetric}). The experimental results are shown in Table \ref{tb:exp1}.
\begin{equation}\label{eq:cmpMetric}
    \sigma = \sqrt{E_{z\in GR_{\textbf{V}}}\left[G(z)-E_{z\in GR_{\textbf{V}}}\left[G(z)\right]\right]}
\end{equation}
We randomly select 10 query samples and compute the average standard deviation of generated sets. Table 1 indicates that our E-GBAS has lower loss (i.e., consistent with the input query) compared to the $\epsilon$-based sampling in all three models and target layers.

\section{Conclusion}
In this study, we propose the explorative algorithm for analyzing the GR to identify generation mechanism in the DGNNs. Especially, we probe the internal layer in the trained DGNNs without additional training by introducing the GB of DGNNs.
To gather samples which satisfy the NRS condition in the complicated and non-convex GR, we applied GB-RRT.
We empirically show that the collected samples in the latent space with the NRS condition share the same generative properties. We also qualitatively indicate that the NRS in the distinct layers implies different generative attributes. 
Furthermore, the concept of the proposed algorithm is general and can also be used to probe the decision boundary in the classifier. So we believe that our method can be extended to different types of deep neural networks.

\section*{Acknowledgement}
This work was supported by the Institute for Information \& communications Technology Planning \& Evaluation (IITP) grant funded by the Ministry of Science and ICT (MSIT), Korea (No. 2017-0-01779, XAI) and 
the National Research Foundation of Korea (NRF) grant funded by the Korea government(MSIT), Korea (NRF-2017R1A1A1A05001456).

{\small
\bibliographystyle{aaai}
\bibliography{main}
}

\end{document}